\documentclass{article}

\usepackage{arxiv}

\usepackage[utf8]{inputenc} 
\usepackage[T1]{fontenc}    
\usepackage{hyperref}       
\usepackage{url}            
\usepackage{booktabs}       
\usepackage{amsfonts}       
\usepackage{nicefrac}       
\usepackage{microtype}      

\usepackage{graphicx}
\usepackage{color}
\usepackage{soul}
\usepackage{bbold}
\soulregister\cite7
\soulregister\ref7
\soulregister\pageref7
\usepackage{algorithm,algorithmic}
\usepackage{amsmath}
\usepackage{subcaption}

\title{DriveGuard: Robustification of Automated Driving Systems with Deep Spatio-Temporal Convolutional Autoencoder}

\author{
  Christos Kyrkou$^1$\thanks{ckyrkou@gmail.com,www.christoskyrkou.com}, Andreas Papachristodoulou$^1$, Theocharis Theocharides$^{1,2}$\\
  \textit{$^1$KIOS Research and Innovation Center of Excellence}\\
  \textit{$^2$Department of Electrical and Computer Engineering}\\
  University of Cyprus\\
  1 Panepistimiou Avenue, Nicosia Cyprus \\
  \{kyrkou.christos,apapac03,ttheocharides\}@ucy.ac.cy \\
}

\begin{document}
\maketitle

\begin{abstract}
Autonomous vehicles increasingly rely on cameras to provide the input for perception and scene understanding and the ability of these models to classify their environment and objects, under adverse conditions and image noise is crucial. When the input is, either unintentionally or through targeted attacks, deteriorated, the reliability of autonomous vehicle is compromised. In order to mitigate such phenomena, we propose \textbf{DriveGuard}, a lightweight spatio-temporal autoencoder, as a solution to robustify the image segmentation process for autonomous vehicles. By first processing camera images with DriveGuard, we offer a more universal solution than having to re-train each perception model with noisy input. We explore the space of different autoencoder architectures and evaluate them on a diverse dataset created with real and synthetic images demonstrating that by exploiting spatio-temporal information combined with multi-component loss we significantly increase robustness against adverse image effects reaching within 5-6\% of that of the original model on clean images. 
\end{abstract}


\section{Introduction}

\begin{figure}[t]
	\centering
	\includegraphics[width=0.7\linewidth]{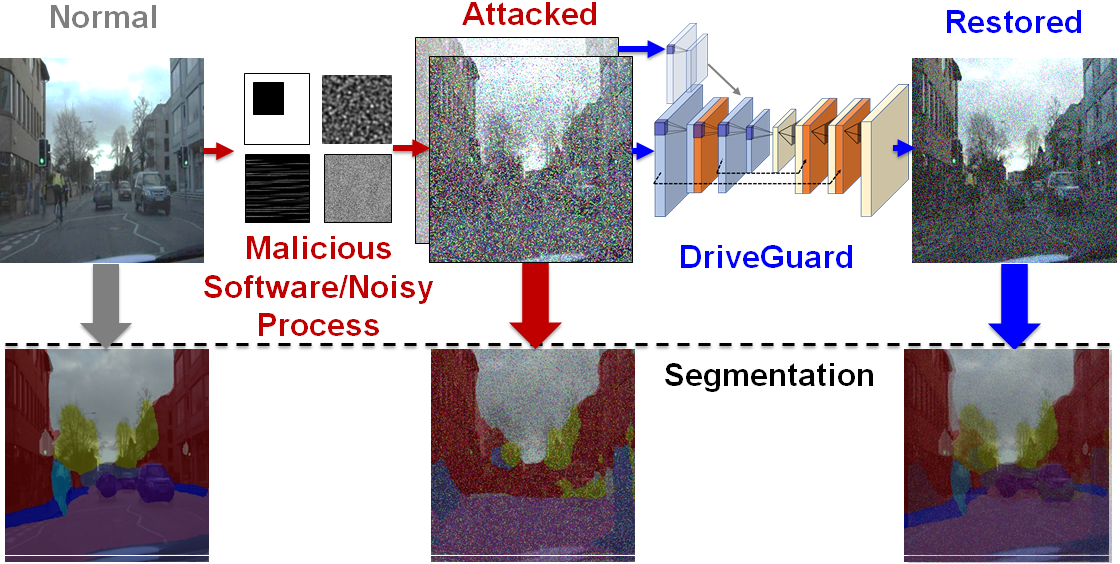}
	\caption{Even simple attacks and image noise models can have a detrimental effect on the image segmetation output. \textbf{Noisy Process:} Effect of noise attack on segmentation output. Notice, that the objects are not localized well and objects such as cyclists are completely lost. \textbf{DriveGuard:} Attempts to restore the image quality resulting in improved segmentation performance.}
	\label{fig:autosensors}
\end{figure}

Automated vehicles are a fundamental component of intelligent transportation systems and will improve safety, traffic efficiency and driving experience, and reduce human errors \cite{CGV-079}. Deep learning solutions are used in several autonomous vehicle subsystems in order to perform perception, sensor fusion, scene analysis, and path planning (e.g., \cite{OD:AD:2019,Reviewer_4_1,Reviewer_4_2,Reviewer_4_3,Reviewer_4_4}). State-of-the-art and human-competitive performance have been achieved by such algorithms on many computer vision tasks related to autonomous vehicles \cite{CarResearch:2019}. Nevertheless, over the last years it was demonstrated that deep-learning-based solutions are susceptible to various threats and vulnerabilities that can cause the autonomous vehicles to misbehave in unexpected and potentially dangerous ways. Besides physical attacks that can induce erroneous behaviour there is also the possibility that the camera data can be manipulated directly thus eliciting false algorithmic inferences. This can cause an AI-based perception module/controller to make incorrect decisions, such as when an autonomous vehicle fails to detect a lane/ parking marking and results in a collision \cite{SecCAV:2020,AdvAD:2020:JAST}.  An example of such an attack can be either data injection by malicious software installation \cite{DandT:2019:Secure} in the CAN bus, projecting patterns \cite{MobilBye:2019}, manipulation of over the air updates \cite{DriveSec:2017,JeepHackers:2014}, or even faults \cite{FDcam:ICCE:2014}. Hence crafting appropriate defenses against attacks that go beyond the hardware layer is important to realize safe autonomous driving \cite{AdvAD:2020:JAST,RemoteAttacks:2015,SecCAV:2020}.

Recently, these systems have attracted increased attention within academia, and the academic community has begun to investigate the systems’ robustness to various attacks targeting the camera sensors and the underlying computer vision algorithms \cite{threat2018}. Even though adversarial attacks have gained much interest in recent years they may fail to manifest in real-world applications \cite{NoNeedAdv:2017}, while studies demonstrated the effect of attacking on the camera sensor data by adding noise and artifacts can also be detrimental \cite{Kyrkou:ISVLSI:2020}. Hence, we focus on the problem of image distortions like added noise and image manipulations, that can cause scene structural elements (e.g. traffic signs, pedestrians, etc.) to not be detected. Such attacks are easy to be added as the attacker does not need to have any information regarding the underlying algorithm and models and can lead to erroneous output of a perception module (Fig. \ref{fig:autosensors}). Existing pre-processing approaches, such as filtering (bilateral or gaussian) even though effective against specific attack types may fail to completely remove artifacts.

This is an extremely challenging task because of its inherent nature that involves both spatial and temporal context. Recently, many deep learning based approaches, such as AutoEncoders are shown to be more effective than these classical methods. However, existing implementations incorporate only on the spatial context \cite{vincent2008extracting}, \cite{mao2016image}. Models that incorporate both Spatial and Temporal Dimensionalities are employed for similar tasks but are based on heavy and inefficient methods, such as LSTMs \cite{patraucean2015spatio}.

To proactively mitigate the effects of deteriorated image quality, in this paper we propose \textbf{\textit{DriveGuard}}, an approach that uses convolutional autoencoder models to improve the robustness of the image segmentation models used in applications of self-driving cars, as shown in Fig. \ref{fig:autosensors}. Note, that while it is possible to train the segmentation model directly with augmentations, our approach allows to simultaneously guard other tasks beyond segmentation. Furthermore, it permits us to also develop suitable mechanisms to detect image quality degradation attacks, however, this goes beyond the scope of this work. Specifically, we investigate the effect of different architectures under different metrics for both image quality and segmentation. Our main contributions are as follows:

• We propose a lightweight spatio-temporal autoencoder, utilizing separable convolutions, as an image reconstruction tool for robustifying semantic segmentation for autonomous vehicle applications.

• We investigate different architectures and loss function combinations, the structural similarity index (SSIM \cite{FDcam:ICCE:2014} and mean square error (MSE), for better structure understanding and pixel-wise restoration respectively. 

• We consider a combination of traditional noise models and more targeted attacks, on a more challenging dataset comprised of realistic and synthetic data with diverse weather conditions generated from an autonomous driving simulator.

We evaluate the mitigation performance of the different approaches using traditional metrics and study its impact on the semantic segmentation outcome. We manage to restore most of the segmentation performance from heavily distorted images to within $5-6\% $ of the original model applied on clean images. Besides the predominance demonstrated by our quantitative results, the incorporation of the temporal context allows our model to mitigate extreme attacks of blank regions and stacked lines which spatial autoencoders were unable to overcome. Overall, this approach provides a more universal defence against different forms malicious processes that can be integrated not only with different segmentation models, but with other 2D tasks as well (e.g., object detection \cite{squeezeDet}) as it is agnostic to the underlying scene understanding model.

\section{Background and Related Work}

\subsection{Problem overview}

Image deterioration attacks aim to alter the input image in order to lead the vehicle perception modules to fail. In contrast to adversarial examples these attacks are not guided by a target label but the objective is to cause a general drop in the image quality so that the perception module’s output becomes erroneous. These attacks are rather simple and do not require an attacker to have knowledge or access to any perception model and thus can be considered as more common. An attacker can manage to access critical vehicle systems and thus be able to directly manipulate the camera image via modification through internal malicious software (Fig. \ref{fig:autosensors}) \cite{SecCAV:2020}. Given such deliberate or unintentional events the objective is to be able to restore the image quality to a good enough level so that the perception module (e.g., semantic segmentation) is able to still perform with adequate accuracy.

\subsection{Image Filtering Techniques}

Methods of image restoration are a topic of research in the last decades \cite{Milanfar:Filtering}. Filtering approaches such as Gaussian Filters used in studies \cite{wand1994kernel}, \cite{8297089}, \cite{mao2016image} where they apply tranformations of pixels based on the information provided by a window of neighbors. The lack of stronger adaptivity to the underlying structure of the image and objects of interest is a major drawback of these classical approaches. Other approaches include the Bilateral Filter \cite{8297089}, \cite{zhang2008multiresolution},  based on the Gaussian Filter, which integrates pixel-wise spatial and photometric disparity. The advantage is that the computational complexity remains low but it employs adaptivity to the window used. On the downside, in low disturbed images it doesn’t perform accordingly. Other works measure frame quality assessments as features, and then the computed features are analyzed for fault detection \cite{FDcam:ICCE:2014}, or try to remove environmental noise such as \cite{ShadowRemove:2020}.

\subsection{Deep Image Restoration}
 
Autoencoders are one type of the profound deep learning methods utilized for the removal of noise in images and restoring image quality \cite{vincent2008extracting}. Denoising auto-encoders can be stacked to form a deep network by feeding output of the previous convolution layer to the current layer as input. Jain and Seung \cite{jain2009natural} proposed to use Convolutional Neural Networks (CNN) to denoise natural images. 
The deployment of fully Convolutional Neural Networks (FCNN) for image denoising was proposed by  \cite{jain2009natural},  \cite{mao2016image}. Such approaches deploy autoencoder architectures similar in structure with the FCNN, commonly used for semantic segmentation or super-resolution \cite{dong2015image}.

This approach is followed by Xiao Jiao Mao \cite{mao2016image}, with a symmetrical architecture between the decoder and encoder layers used for image restoration. It aims to reconstruct perturbed images by upsampling the features encoded in the latent space by the encoder and thus recovering degraded features from the image. The contribution of this approach is the deployment of skip-layer connections, which strengthen the autoencoder by feeding lower-level features from early layers to later layers in the network. It has also the advantage of the training converging faster.

The effect of different loss functions was tested in the tasks of Image restoration and super-resolution by \cite{8297089}, \cite{zhao2016loss}. Zhao et al,\cite{zhao2016loss} proposed the application of SSIM as a loss function in the sense that it would provide a better understanding of the structure in the images. They compare the de-facto applied squared norm of error loss with the loss functions of the SSIM family on both forementioned tasks. They conclude that their proposals outperform the models trained by the standard squared error not only due to the better structure understanding but also by its poor convergence performance. The same approach of replacing pixel-wise loss functions such as MSE with structure similarity ones was followed by\cite{8297089}, in the task of super-resolution.

The idea of a spatio-temporal autoencoder was investigated in different tasks and approaches by \cite{patraucean2015spatio}, \cite{8449842}.
 In the case of \cite{patraucean2015spatio} the aim was the prediction of the next frame with the application of differentiable memory cells for the task of path-planning, action recognition etc. 
In \cite{8449842} the authors looked into the application of video deblurring purposes. In order to incorporate both the spatial and temporal dimensionalities they deployed three dimensional convolution in both domains. Their approach involved the encoding of multiple neighboring frames. The model was further extended where it was trained using adversarial training where the adversarial loss was established to fool the discriminator on misidentified generated images as real life ones.
A similar approach of a CNN model handling spatio-temporal information was followed by \cite{Su_2017_CVPR}, with the task of recovering the central of the input frames. 
In \cite{8449842}, neighboring frames are stacked as the temporal dimensionality of the data, whereas in \cite{Su_2017_CVPR} they deploy 2D convolutions to handle the time axis as an extension to the spatial RGB dimension.

In these works the main focus was on reconstructing natural images. Yet the utilization of such methods for autonomous mobility applications have not been investigated. Considering the importance of robustifying autonomous vehicles against different types of noise and attacks, we explore the appealing potential of image reconstruction using deep convolutional autoencoders.

\section{Proposed Approach}

\subsection{Frame Quality Degradation}

To simulate image quality degradation events we explore different artificial deterioration techniques which include the addition of noise to the sensed data (Gaussian, Salt and Pepper, Poisson, etc.), adding artifacts such as horizontal/vertical lines, color blanking out image regions. Then metrics such as the reconstruction error can be used as a means for evaluation of the approach. Moreover, the artificial noise methods are required to be representative of real-life image attacks. These methods include approaches that follow probabilistic distribution models such as random Gaussian as well as targeted attacks such as the scenario of selective manipulation of the input image like the addition of artifacts or blank Regions \cite{shetty_neurips2018}. 

\subsubsection{Traditional Degradation Techniques}

The first batch of applied noise distortions concern the following universally applied and standard noise methods: Random Gaussian \cite{GuoBlind2018}, Speckle\cite{SpeckleNoise2015}, Salt and Pepper, Poisson \cite{PracticalPoissonian-Gaussian} as well as combinations of these \cite{TeslaAttack:2019}. The noise level of every noise method is parametrized and probabilistically applied using \cite{van2014scikit}. In addition, various combinations of noises are also possible which can render the use of a single technique less successful, hence the need for a machine-learning-driven approach.

\noindent$\bullet$ \textbf{Gaussian Noise} that disturbs the intensity of the pixel $x$ to arrive at the modified intensity $g(x) = \frac{1}{\sigma\sqrt{2\pi}}\exp^{\frac{-(x-\mu)^2}{2\sigma^2}}$, where $\sigma$ is the standard deviation of the distribution and $\mu$ is the mean.

\noindent$\bullet$ \textbf{Speckle Noise} is multiplicative noise such that $g(x)=x+n\times x$, where $n$ is a random noise model such as gaussian.

\noindent$\bullet$ \textbf{Salt and Pepper Noise} noise is applied by replacing a proportion of pixels with noise on range [min\_dark\_value, max\_bright\_value].

\noindent$\bullet$ \textbf{Poisson Noise} is noise generated from the data such that the corresponding output pixel $g(x)$ uses a Poisson random number generated from a Poisson distribution, with mean that of pixel $x$.

\subsubsection{Artifact Noise}

Another possible attack scenario is the selective addition of artifacts over specific image regions that can potentially fool the scene perception module. Going beyond traditional noise models we aim on mimicking such attacks by adding blank regions and lines on random locations of the image (e.g., Fig. \ref{fig:autoencoders}). Blank Regions and stacked lines represent the purposely occluded regions and stacked pixels deployed by the attacker as well as possible dirt-particles that might reproduce such phenomenon on the camera. These artifacts are added in random locations in the image and their number and size can also vary. 

\subsection{Deep Learning Architecture}

\begin{figure}[t]
	\centering
	\includegraphics[width=0.7\linewidth]{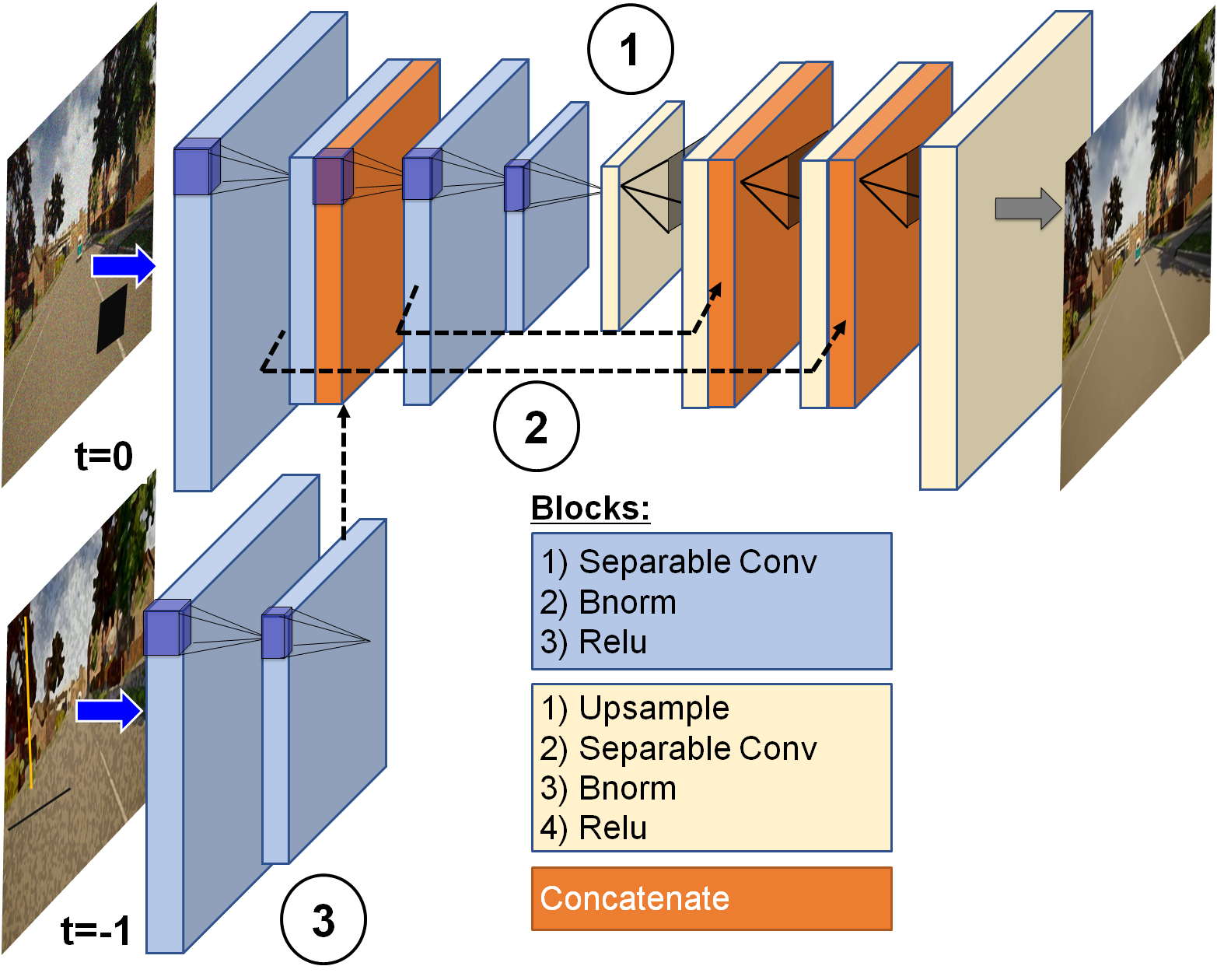}
	\caption{Several autoencoder architectures are investigated: 1) \textit{AE}: Autoencoder with downsampling and upsampling path. 2) \textit{SCAE}: Autoencoder with skip connections for feature enhancement. 3) \textit{STAE}: A spatio-temporal autoencoder with a path that also extracts features from the previous frame. (Best viewed in color)}
	\label{fig:autoencoders}
\end{figure}

We utilize the convolutional autoencoder family of deep learning networks to build our approach for efficient image reconstruction. We explore different models as shown in Fig. \ref{fig:autoencoders}, formulate a loss to train a model to reconstruct a degraded image. In all cases the networks are composed of an encoder that finds an efficient smaller representation for the latent space and a decoder that upsamples the latent features in order to reconstruct it to the original dimension. The model is trained to minimize the reconstruction error based on unsupervised training with the unaltered input image as the ground truth. For the design and implementation of our autoencoder we followed a progressive approach of gradually refining the model with additional techniques and the outcome of each refinement is further analyzed in the evaluation section. 

\subsubsection{Traditional Autoencoder (AE)}

The first model is a denoising autoencoder that consists of $4$ Layers of encoding, and $4$ layers of decoding. It follows a symmetric encoding-decoding relationship, where corresponding layers in the encoder-decoder have the same depth and dimensions.

For the encoder, depthwise separable 2D convolutions are used instead of classical 2D convolutions to lower the number of parameters and make the computation more efficient. The depthwise separable convolution deals with both the spatial dimensions, and the channel dimensionality as well. It factorizes the main convolution operation into depthwise and pointwise convolutions. Following every separable convolution we applied \textit{BatchNormalization} at every Layer and used a \textit{ReLU} as the activation function as shown in Fig. \ref{fig:autoencoders}. 

\subsubsection{Autoencoder with Skip Connections (SCAE)}

Early layers are able to identify the lower level features that mainly depict the latent structure. However, the reduction of spatial resolution can result in fine level details being lost during the downsampling process. This results in poor structure understanding from the decoder which is unable to reconstruct. To recover them during the upsampling process we introduce skip connections in the autoencoder architecture. Skip connections in autoencoder architectures, bypass a number of layers in the network and are combined with an output of a more advanced layer. In addition, the latter layers are fed with those fused features. Since we are focused on retaining the semantic understanding of an image and not solely concentrated on denoising the image, such features are of higher importance. An additional advantage is that it allows better gradient flow, since the loss gradients are values less than 1, the backpropagating multiplication of successive layers result in insignificant gradient values at the first layers. We added two symmetric skip connections where lower level features are fed to the decoder layers and are combined through concatenation along the channel axis. In our autoencoder architecture the output features of the first and third layers of the encoder were connected to the second and third layers of the decoder respectively.  

\subsubsection{Spatio-Temporal Autoencoder (STAE)}

Intuitively simple architectures that receive single frames as inputs have the major weakness of basing their understanding only on the spatial dimensionality. On the other hand, humans don’t base their understanding of structure or higher-level semantics from single input snaps-frames of their environment. They make predictions not only from appearance and the spatial-axis, but also by motion and the temporal axis \cite{patraucean2015spatio}.  In order to incorporate the temporal dimension in our approach we also encode information from preceding frames. Through the deployment of sequences of frames as input to the model we strengthen its understanding of object structure. It is an effective approach since autoencoders are shown to be capable of aligning information from consecutive frames where the motion of objects as well as shifts in the background are handled in techniques of video deblurring and 3D reconstruction \cite{patraucean2015spatio}, \cite{1222588}. Furthermore, we expect that the sequence approach can be effective on identifying noise inconsistencies through the temporal dimension and that perturbed features can be reconstructed through accumulation from consecutive frames. In addition, the utilization of a multi-frame autoencoder avoids the deployment of LSTMs GRUs etc, which makes the whole process more efficient.

This version of our autoencoder incorporates a second input stream which encodes the information of the previous frame in parallel with the current-frame stream, for the first two layers. The two streams are concatenated and inputted to the third encoding layer. The rest of the autoencoder structure and its skip connections remain the same, thus, the skip connection that passes information from the output of the first encoding layer to the third decoding layer is not depreciated. The reason behind this is that we aim for our structure to be localised as in the current frame, thus, we pass lower-level features only from the current-frame stream. Furthermore, the joint output of the third encoding layer still feeds the second decoding layer.

\subsection{Loss function}
To train the different autoencoder architectures we explore two loss functions which we apply as standalone and combination to study their influence. Specifically, we adopt the mean square error (MSE) which is typically used to directly regress pixel values, and the structural similarity index measure (SSIM) which can be used to assess the structural similarity between two images. We choose these among the plethora of existing indexes, because they are established measures, and because they are differentiable which is requirement for the backpropagation stage. 

• The Mean Square Error (MSE) loss is widely used in optimization objectives and for restoring image quality in many existing methods. For a given size $m\times n$ image $I$ and noise image $K$, the mean square error is defined as:

\begin{align}
    L_{MSE} = \dfrac{1}{mn}\sum_{i=0}^{m-1}\sum_{j=0}^{n-1}[I(i,j)-K(i,j)]^2 .
\end{align}

• SSIM is an index to compare the similarity of two image from the perspective of brightness, contrast, and structure. The definition of SSIM is given below and is based on a window by window basis:

\begin{align}
    SSIM = \dfrac{(2\mu_I\mu_K+c_1)(2\sigma_{IK}+c_2)}{(\mu_I^2+\mu_K^2+c_1)(\sigma_I^2+\sigma_K^2+c_2)},
\end{align}

\noindent where $\mu_I$ is the mean of the window from image $I$,$\mu_K$ is the mean of the window from image $K$, $sigma_I^2$ is the variance of the window from image $I$, $\sigma_K^2$ is the variance of the window from image $K$, and $c_1,c_2$ are constants set to their default value as in \cite{SSIM2004}. Subsequently, the SSIM-based loss is the average of the SSIM value across window with dimension $(2N+1)$ centered at $(i,j)$ and across image channels. 

\begin{align}
    L_{SSIM} = \dfrac{1}{mn}\sum_{i=N}^{m-N}\sum_{j=N}^{n-N}SSIM(I(i,j),K(i,j)).
\end{align}

The two losses are combined as follows to form a single loss $L = \lambda_{MSE}\times L_{MSE} + \lambda_{SSIM}\times (1-L_{SSIM})$, where $\lambda_{MSE}=1$ and $\lambda_{SSIM}=0.1$ are empirically selected scaling factors used to balance the loss components. We train models with each loss individually and combined to further understand their impact. 

\subsection{Training}
To train the different autoencoder models we construct image pairs $(x,x')$ where $x'$ is the input image, which may be distorted and $x$ is the undistorted/clean image version. The model is uninformed of the attack; therefore it must preserve the image contents and object structures in the case which the image is not distorted. In order to achieve this, we needed to use a ratio of undistorted/clean images for both the training and evaluation stages. Moreover, each noise degradation method is applied stochastically to generate $x'$. In addition, we also apply the same geometric transforms (cropping, rotation, zooming) and photo transformations (HSV jitter, gamma transforms, etc.) to each image pair as additional augmentations. Overall, we maintain $1$ clean image in every $4$ frames. The overall, approaches are implemented in Tensorflow/Keras and each model is trained for $200$ epochs, with learning rate of $1\times 10^{-4}$.

\section{Experimental Methodology and Evaluation}

In this section we describe the general evaluation and data generation processes as well as the metrics used for the performance evaluation. An important aspect in our methodology is that we use the performance of an already established semantic segmentation model, specifically DeepLabv3 \cite{DeepLabV3} trained using the Cityscapes dataset, as our baseline. In this way we study how the proposed approach can preserve the semantic information of synthetic images and the original model performance. Deeplabv3 achieves a 78.6 IoU evaluation score on the Cityscapes dataset without applying image degradation augmentations. Hence, with the introduction of noise and the more diverse dataset we expect this performance to be reduced. 

\subsection{Datasets}

\textbf{Realistic Dataset:} The Cityscapes dataset (\textbf{\textit{CS}}) \cite{Cordts2016Cityscapes} is our choice for real-life data for the training and evaluation phase. It provides the associated high-quality semantic segmentation annotations with 3900 and 880 images for train and test respectively, annotated at 20 frames per second from 50 different cities. It is commonly used as the data source for autonomous driving computer vision tasks and state-of-the-art deep learning models.

\textbf{Diverse conditions - synthetic dataset:} Besides artificially deployed distortions, the model needs to be robust against different environmental conditions which include different regional sceneries and different weather conditions. 
We employ the CARLA simulator \cite{CARLAsimevasion:2019} to construct a synthetic dataset (\textbf{\textit{CL}}). 
We form a dynamic weather challenge to evaluate our model’s noise mitigation performance in conditions where the image reconstruction is more difficult. We deploy an autonomous vehicle agent to collect RGB image data as well as ground truth semantic segmentation maps from 5 different towns with dynamic weather conditions and realistic graphical settings (4 towns used as training images (1800), 1 town as test images (490) and 1 town as validation images (450)).
Diverse weather conditions include cloudy; rainy weather as well as sunset and night light. The data are continuous and are recorded in high frame rate so that they can be used both as unique instances as well as sequences to exploit the temporal domain. It is important to note that on synthetic data of clear and cloudy weather conditions at noon time, the Deeplabv3 model faces an IoU score drop of only about  $12.5\%$, while on dynamic weather data which cover $80\%$ of the synthetic dataset, a further drop of about $24\%$ is observed. 

A \textbf{joined dataset} was used for the training and testing of both the autoencoder and semantic segmentation model. Cityscapes semantic segmentations are labelled based on 30 classes whereas Carla labels semantic segmentations using only 13 classes. Thus, we devised a common classification format and alignment between semantic segmentation classes for the two data sources. 

\subsection{Quantitative Evaluation Results}

The evaluation process involved examining the different trained models and configurations under different levels of deterioration attacks and observe how well each can reconstruct the image and also robustify the segmentation process. To evaluate the effectiveness of the different approaches on mitigating the adverse effects of deterioration attacks on the semantic segmentation output, we utilize semantic segmentation metrics on the diverse synthetic data (CL), which provides access to sequential ground truth labels. Beyond that we also use peak signal to noise ratio (PSNR) along with MSE and SSIM as metrics for evaluating the actual reconstruction performance on the joined dataset (CS \& CL). These are summarized below:

\textbf{Pixel Accuracy:} Percentage of pixels in the image that are classified correctly. 

\textbf{Intersection-Over-Union Score:} the area of overlap between the predicted segmentation and the ground truth divided by the area of union between the predicted segmentation and the ground truth.

\textbf{PSNR:} The ratio between the maximum possible power of a signal and the power of corrupting noise defined by MSE and given by $10\times log_{10}\frac{MAX_I^2}{MSE}$

Next we evaluate how each of the refinements have contributed in improving the robustness of image segmentation. To do so we evaluate 5 different autoencoder models using different loss combinations, 2 traditional image processing denoising filters (median and bilateral), and an approach with unguarded input as baseline. We tested each method with 5 different noise levels comprising different parameter values for each degradation method (Gaussian, Speckle, Salt and Pepper, and Poisson). Along with these Noise Methods the input image was degraded through the addition of artifacts, of which the number was selected based on the noise level. The parameters used are shown in Table \ref{tab:noiseparams}.

\begin{table}[]
\centering
\resizebox{0.6\columnwidth}{!}{
\begin{tabular}{|l|c|c|c|c|c|}
\hline
\textbf{Noise Level}  & \multicolumn{1}{c|}{0 - Clean} & \multicolumn{1}{c|}{1} & \multicolumn{1}{c|}{2} & \multicolumn{1}{c|}{3} & \multicolumn{1}{c|}{4} \\ \hline
\textbf{Amount (s\&p)}    & 0    & 0.1      & 0.2   & 0.3   & 0.4 \\ \hline
$\boldsymbol{\sigma}^2$ & 0   & 1  &  4  & 9 & 16  \\ \hline
\textbf{Artifacts - Range}&  0  &  1 - 13   & 13 - 25  & 25 - 37& 37 - 49\\ \hline
\end{tabular}
}
\caption{Parameters for each noise level.}
\label{tab:noiseparams}

\end{table}

\begin{table*}[]
\centering
\resizebox{\textwidth}{!}{%
\begin{tabular}{lllllllll}
\cline{1-8}
\multicolumn{1}{|l|}{} &
  \multicolumn{1}{l|}{\textbf{SCAE(SSIM-MSE)}} &
  \multicolumn{1}{l|}{\textbf{STAE(SSIM-MSE)}} &
  \multicolumn{1}{l|}{\textbf{AE(MSE)}} &
  \multicolumn{1}{l|}{\textbf{SCAE(MSE)}} &
  \multicolumn{1}{l|}{\textbf{SCAE(SSIM)}} &
  \multicolumn{1}{l|}{\textbf{Bilateral Denoising}} &
  \multicolumn{1}{l|}{\textbf{Median Filtering}} &
  \textbf{} \\ \cline{1-8}
\multicolumn{1}{|c|}{\textbf{MSE}} &
  \multicolumn{1}{c|}{0.001609809} &
  \multicolumn{1}{c|}{\textbf{0.001590869}} &
  \multicolumn{1}{c|}{0.002378508} &
  \multicolumn{1}{c|}{0.001650406} &
  \multicolumn{1}{c|}{0.001912629} &
  \multicolumn{1}{c|}{0.027107959} &
  \multicolumn{1}{c|}{0.002277899} &
  \multicolumn{1}{c}{} \\ \cline{1-8}
\multicolumn{1}{|c|}{\textbf{PSNR}} &
  \multicolumn{1}{c|}{28.88083292} &
  \multicolumn{1}{c|}{\textbf{29.01536878}} &
  \multicolumn{1}{c|}{26.85623264} &
  \multicolumn{1}{c|}{28.64276771} &
  \multicolumn{1}{c|}{28.2089766} &
  \multicolumn{1}{c|}{25.35342745} &
  \multicolumn{1}{c|}{28.1580106} &
  \multicolumn{1}{c}{} \\ \cline{1-8}
\multicolumn{1}{|c|}{\textbf{SSIM}} &
  \multicolumn{1}{c|}{0.809236498} &
  \multicolumn{1}{c|}{\textbf{0.834861451}} &
  \multicolumn{1}{c|}{0.741865986} &
  \multicolumn{1}{c|}{0.786420672} &
  \multicolumn{1}{c|}{0.816084715} &
  \multicolumn{1}{c|}{0.488412969} &
  \multicolumn{1}{c|}{0.747806808} &
  \multicolumn{1}{c}{} \\ \cline{1-8}
\multicolumn{1}{|c|}{\textbf{Pixel Accuracy$^*$}} &
  \multicolumn{1}{c|}{0.614963399} &
  \multicolumn{1}{c|}{\textbf{0.679464035}} &
  \multicolumn{1}{c|}{0.529475664} &
  \multicolumn{1}{c|}{0.570991854} &
  \multicolumn{1}{c|}{0.611822439} &
  \multicolumn{1}{c|}{0.545451794} &
  \multicolumn{1}{c|}{0.5278731} &
  \multicolumn{1}{c}{\textbf{}} \\ \cline{1-8}
\multicolumn{1}{|c|}{\textbf{IoU Score$^*$}} &
  \multicolumn{1}{c|}{0.262517246} &
  \multicolumn{1}{c|}{\textbf{0.302267252}} &
  \multicolumn{1}{c|}{0.20174148} &
  \multicolumn{1}{c|}{0.243014075} &
  \multicolumn{1}{c|}{0.256787356} &
  \multicolumn{1}{c|}{0.199243602} &
  \multicolumn{1}{c|}{0.248135077} &
  \multicolumn{1}{c}{\textbf{}} \\ \cline{1-8}

\end{tabular}%
}
\caption{Average metrics across the different noise levels for different models using the joined dataset (CS \& CL).$^*$Segmentation metrics evaluated on CL.}
\label{tab:avgmetrics}
\end{table*}

\begin{figure}[t]
	\centering
	\includegraphics[width=0.7\linewidth]{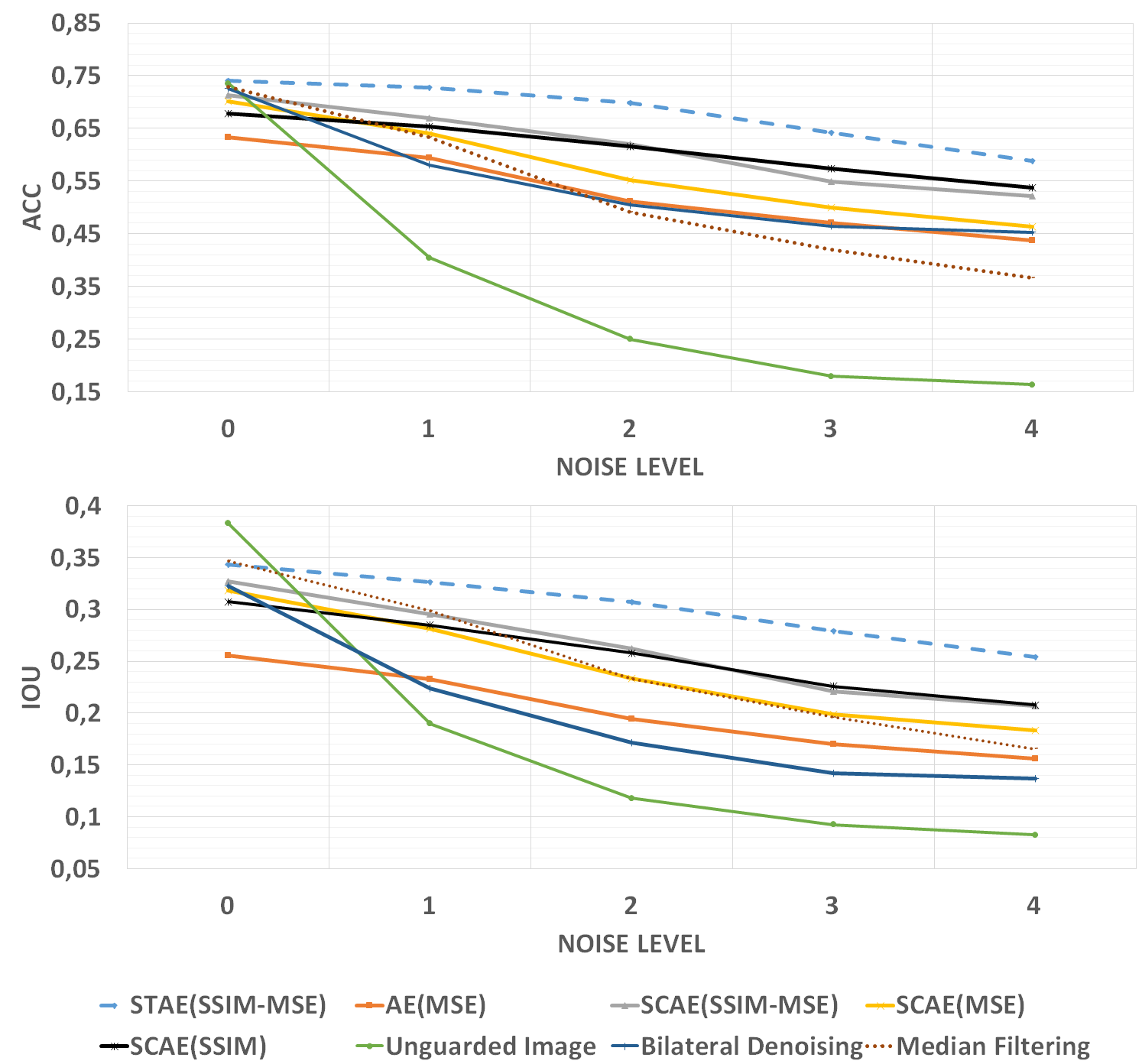}
	\caption{Semantic segmentation metrics for different models/methods under different noise levels on the CL dataset.}
	\label{fig:semsegres}
\end{figure}

\begin{figure}[t]
	\centering
	\includegraphics[width=0.8\linewidth]{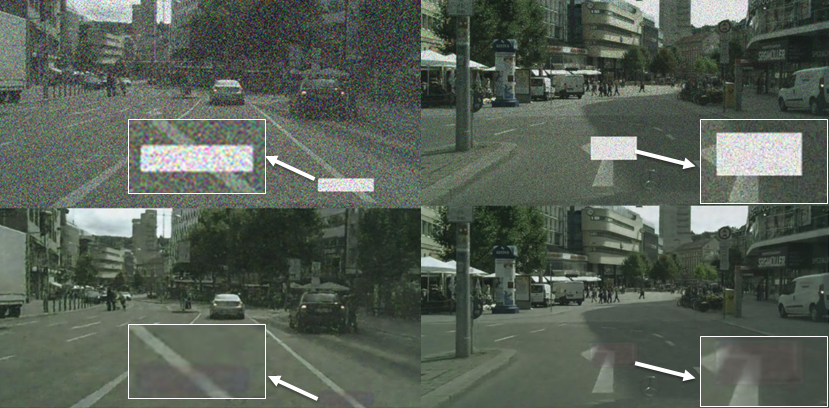}
	\caption{Results on CS Dataset: The autoencoder manages to remove the noise and utilize the temporal aspect to restore the missing line and arrow on the road.}
	\label{fig:resblank}
	\vspace{-10pt}
\end{figure}

\begin{figure}[t]
	\centering
	\begin{subfigure}[b]{0.8\columnwidth}
		\centering
		\includegraphics[width=\textwidth]{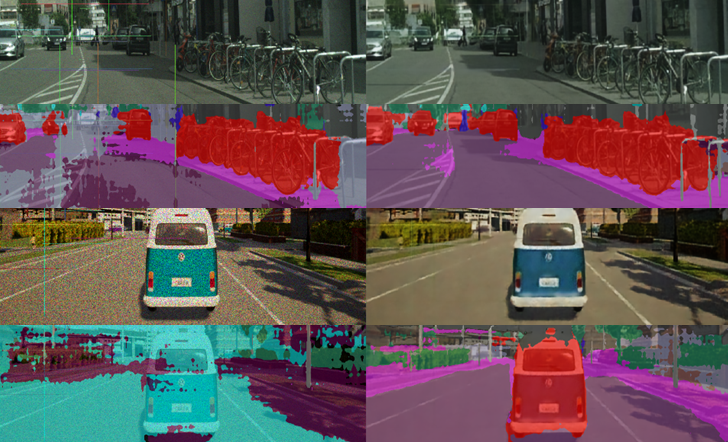}
		\caption{\textit{left}: Attacked, \textit{right}: Restored}
		\label{fig:res_scenes}
	\end{subfigure}
	\\
	\begin{subfigure}[b]{0.8\columnwidth}
		\centering
		\includegraphics[width=\textwidth]{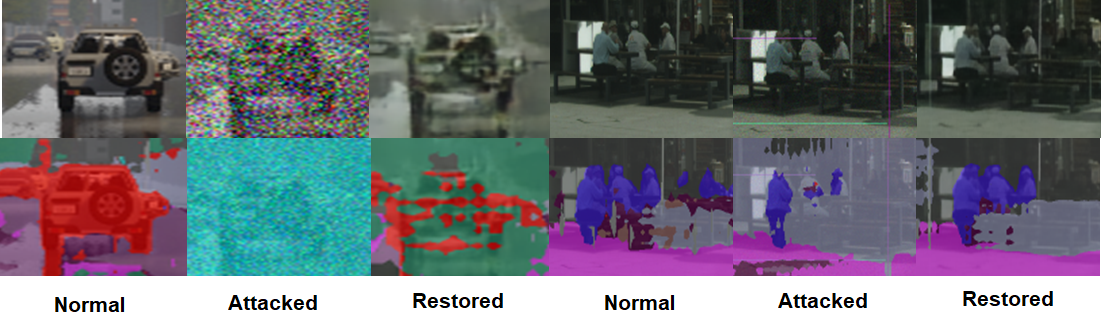}
		\caption{\textit{left}: Normal, \textit{center}: Attacked, \textit{right}: Restored}
		\label{fig:res_obj}
	\end{subfigure}
	\caption{Semantic segmentation results on CS and CL data on normal, attacked, and restored images. The autoencoder manages to restore the semantic segmentation for important objects.}
	\label{fig:semseg_images}
	\vspace{-10pt}
\end{figure}

\subsubsection{Impact of Skip Connections}

The simple-plain autoencoder trained using MSE loss can effectively reduce noise from an image and make the scene perception process more robust as it achieved to double the semantic segmentation metrics on attacked images. However, this simple autoencoder had comparable quantitative results with the traditional approaches. Specifically, the autoencoder could not retain the quality of clean images, and was outperformed by both Bilateral and Median methods, something observed through both the quantitative and qualitative results. However, in noise levels of higher complexity the model had more consistent performance on pixel accuracy and IoU score.

In terms of the comparison between the autoencoders, the refined autoencoder with skip connections is significantly stronger against all the different noise levels compared to the plain autoencoder and the traditional approaches. When evaluated using the traditional metrics it manages to have $30\%$ better mean square error and better structure similarity by about $4.5\%$ than the plain autoencoder as shown in Table \ref{tab:avgmetrics}. When evaluated using the perception model metrics the difference in improvement remains clear but not as significant. The same is observed from visual results where the increase in mitigating performance suggested by the pixel-wise metrics is not as significant. The major visual improvements are in terms of the retained structure, a result of the lower-level features passed forward through the skip connections.  

An important observation that solidifies our earlier intuitions on the matter is that the perception engine’s evaluation trends are not analogous to the mitigation trends of the pixel-wise metrics. In contrast, SSIM has a more analogous relationship with the perception’s engines performance. Thus, the choice of SSIM as a loss function seems more appropriate in order to train an autoencoder which can make the perception process more robust.

\subsubsection{Impact of Loss Functions}

It is also important to compare the effect of the loss functions used for training on the subsequent performance. We train the skip connection autoencoder with MSE loss function, SSIM loss function and a combination of the two. We cannot deduce any inferences from the MSE, PSNR and SSIM metrics due to their direct or indirect use in the training process. In both metrics used in the semantic segmentation evaluation however, we observe some trends. When the image is clean the model trained under MSE loss retains the image quality better whereas under low levels of noise the two models perform similarly. As the noise attacks become stronger the model trained using SSIM loss is more robust.

Interestingly, our intuitions followed in the design process are justified since there is a clear trade-off between the two loss functions. MSE is a pixel-wise metric, thus, can learn to retain and mitigate smaller details than the regional SSIM. However, as the attacks become stronger and the perturbations become regional they cannot be recovered using pixel-wise information, where the SSIM is a more robust approach. As shown from Fig. \ref{fig:semsegres} a combination of the two Loss functions is able to provide the benefits of both. 

We observe the same trends by both metrics of the semantic segmentation as in the comparison of the MSE and SSIM loss functions. Our approach on combining the two loss functions is clearly effective as their combination provides in general, the best results among them in terms of segmentation. The model trained under the combination of the two loss functions outperforms the earlier versions in retaining the image quality of clean images as well as mitigating noise in the case of low and medium levels of attacks. In the case of high and extreme levels of attack it performs similarly to the model trained using SSIM. Visually, the model is significantly more effective on retaining the image quality of clean images as well as mitigating low to medium levels of attacks than the previous versions.

\subsubsection{Impact of Spatio-Temporal Information} 

Based on the results, it is evident that the sequence-based spatio-temporal autoencoder outperforms the simpler models under every evaluation metric across all noise levels as the average value over the noise levels is the highest. Interestingly, the model outperforms in traditional metrics even the models that used the equivalent metric as their loss-function. In respect to the patch-level metric (SSIM), the results prove that the model can appropriately align features from sequences of frames and handles shifts in motion. The effective accumulation of features from the different frames is evident by the pixel-wise metrics (PSNR and MSE).  

In terms of the perception engine evaluation, the spatio-temporal autoencoder significantly outperforms the rest of the autoencoders. The most interesting observation however is that the model manages to reach pixel accuracy equivalent to the unguarded image when faced with a non-attacked image, at $74.9\%$. Moreover, it keeps its performance above $70\%$ for attacks which are not extreme, and above $60\%$ when the noise is extreme, which appealingly is the achievable performance of the plain autoencoder when presented with a clean image. In terms of Intersection-over-Union score, it faces minor loss in comparison with a clean unguarded image, but manages to have a robust performance under extreme attacks where it performs equivalently to the SSIM trained model for clean images.

\subsection{Qualitative Evaluation Results} 

As shown in the visual results in Fig. \ref{fig:res_scenes} the proposed spatio-temporal approach manages to recover information from the attacked image and restore the segmentation performance. Interestingly even under mild noise artifacts the segmentation map may be highly affected, whereas the ST-AE manages to mitigate these adverse effects. As a result, important elements such as other vehicles(Fig. \ref{fig:res_obj}), pedestrians(Fig. \ref{fig:res_obj}), and drive-able area(Fig. \ref{fig:res_scenes}) are correctly identified. As shown in the visual results in Fig. \ref{fig:resblank}, the Spatio-Temporal autoencoder can also recover sections of the image which were occluded by artifacts (e.g., blank regions). On the other hand, even if all the spatial-only models were trained to counter such attacks, none was effective on overcoming it. This validates our expectation that its multi-frame input nature can accumulate features between inputs to recover perturbations. The results are extremely promising as they establish the model’s robustness against attacks that involve selective occlusion of important features in the frame such as road-lines. The model can uncover important features and localise them appropriately while at the same time denoising the image effectively.

\section{Conclusion}
Future autonomous vehicles are heavily dependent on computer vision and artificial intelligence techniques to perceive their environment. To ensure robust and safe operation these systems need to be safeguarded against attacks (malicious or otherwise). To this end, this paper has demonstrated \textbf{\textit{DriveGuard}} as an approach for robustifying semantic segmentation models, commonly employed in autonomous vehicles, against various forms of image deterioration attacks. In particular, through an exploration of deep autoencoder architectures we have come up with a lightweight spatio-temporal autoencoder that manages to robustify semantic segmentation models.

\section*{Acknowledgement}
This work was supported by the European Union’s Horizon 2020 research and innovation programme under grant agreement No. 833611 (CARAMEL). This work was also supported by the European Union's Horizon 2020 research and innovation programme under grant agreement No 739551 (KIOS CoE) and from the Government of the Republic of Cyprus through the Directorate General for European Programmes, Coordination and Development.

\bibliographystyle{unsrt}
\bibliography{references}

\end{document}